\documentclass{article}
\usepackage{spconf,amsmath,graphicx}
\usepackage{CJKutf8}

\title{External Knowledge Augmented Polyphone Disambiguation Using Large Language Model}
\name{Chen Li}
\address{Ant Group}
\begin{document}
%
\maketitle
\begin{abstract}
One of the key issues in Mandarin Chinese text-to-speech (TTS) systems is polyphone disambiguation when doing grapheme-to-phoneme (G2P) conversion. In this paper, we introduce a novel method to solve the problem as a generation task. Following the trending research of large language models (LLM) and prompt learning, the proposed method consists of three modules. Retrieval module incorporates external knowledge which is a multi-level semantic dictionary of Chinese polyphonic characters to format the sentence into a prompt. Generation module adopts the decoder-only Transformer architecture to induce the target text. Postprocess module corrects the generated text into a valid result if needed. Experimental results show that our method outperforms the existing methods on a public dataset called CPP. We also empirically study the impacts of different templates of the prompt, different sizes of training data, and whether to incorporate external knowledge.
\end{abstract}
\section{Introduction}
\label{sec:intro}

\begin{figure}
    \centering
    \includegraphics[width=0.8 \linewidth]{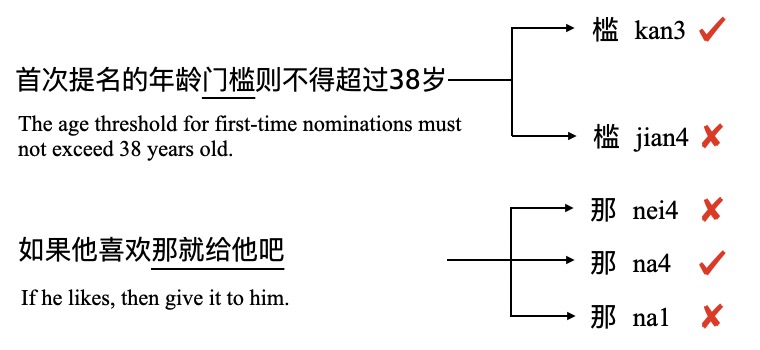}
    \caption{Examples of polyphone in different contexts}
    \label{fig:f1}
\end{figure}

Grapheme-to-phoneme (G2P) conversion plays an important role in the front end of text-to-speech (TTS) synthesis systems, which is mainly a text-processing module. The target of G2P in Mandarin Chinese TTS is to convert Chinese characters into pinyin which is the pronunciation system of Chinese in the form of the Latin alphabet. However, as a symbolic language, some Chinese characters may have more than one pronunciation, and they are often called polyphonic characters. Therefore, it is necessary to do polyphone disambiguation before synthesizing speech, and usually, collocation and context information around the characters would be leveraged. Figure \ref{fig:f1} shows some examples of this phenomenon. Generally, a polyphonic character has a commonly used pinyin and a few rare pinyin.

There has already been a lot of research focused on the problem of polyphone disambiguation. In early publications, some rule-based approaches were proposed, which heavily relied on the pronunciation dictionary and pre-defined rules. For example, utilizing the frequency of different pronunciation and context patterns, which can be mined from a large corpus, is quite straightforward \cite{zhang2001disambiguation}. However, the construction of hand-crafted rules is costly and sometimes inaccurate. Some researchers further build a decision list for polyphonic characters through collocation distribution in a context where they appear. Based on the decision list, the likelihood of a pinyin in a certain collocation can be calculated \cite{dong2004grapheme}. Later, some statistical methods achieve good benchmarks, like Decision Trees (DT) and Maximum Entropy (ME) models \cite{You2011PolyphoneDB,2012Polyphone}.

Recently, with the advancement of Neural Networks (NN), a lot of studies have been applying NN models to polyphone disambiguation. Most of the research describes the problem as a classification task, such as adopting a bi-directional long-short-term memory (BLSTM) layer to encode the input sentence and an output layer for producing probability distribution over all candidate pronunciations \cite{rao2015grapheme,shan2016bi}. Furthermore, a few studies show that taking advantage of multi-level embedding features is helpful, including characters, words, and their positions \cite{cai2019polyphone,dai2019disambiguation}. There is also a series of research that defines the G2P conversion as a Machine Translation (MT) task. A paradigm of encoder-decoder generation is employed in the task to train a Seq2Seq model with attention mechanism \cite{yao2015sequence,zhang2020distant}.

The existing methods are suffering from a lack of supervised data and deep semantic features. As the trends of large language models (LLM) grow, a lot of tasks manage to follow the generation paradigm to acquire stronger generalizability and flexibility. GPT \cite{radford2018improving,brown2020language} uses a massive amount of unlabeled text to pre-train a multi-layer Transformer decoder which shows better results on both natural language understanding and generation tasks. Based on autoregressive models, GLM \cite{du2021glm} improves blank filling pretraining by 2D positional encodings and supports different types of tasks by varying the number and lengths of blanks.

Inspired by LLM, we propose an approach to solving the problem by language models and external knowledge. Specifically, we consider polyphone disambiguation as a generation task that takes the sentence and essential information as model input and directly generates the target pinyin. As known by previous research, the performance of LLM heavily relies on the input prompt, so we also augment it by adding external knowledge such as a semantic dictionary. It can be easily crawled from the Internet and it would be restructured to a multi-level knowledge base for retrieval. Unlike the past methods, our design can also predict unseen characters without any incremental training only if they are in the dictionary. The contribution of this paper can be summarized as follows:
\begin{enumerate}
    \item The proposed method is the first work to introduce LLM techniques into polyphone disambiguation.
    \item We utilize a multi-level semantic dictionary as external knowledge for better performance as well as the adaption to unseen characters.
    \item With the external knowledge, the proposed method outperforms the baseline models in accuracy on the CPP dataset.
\end{enumerate} 

\section{Method}
\label{sec:method}
\begin{figure}
    \centering
    \includegraphics[width=1\linewidth]{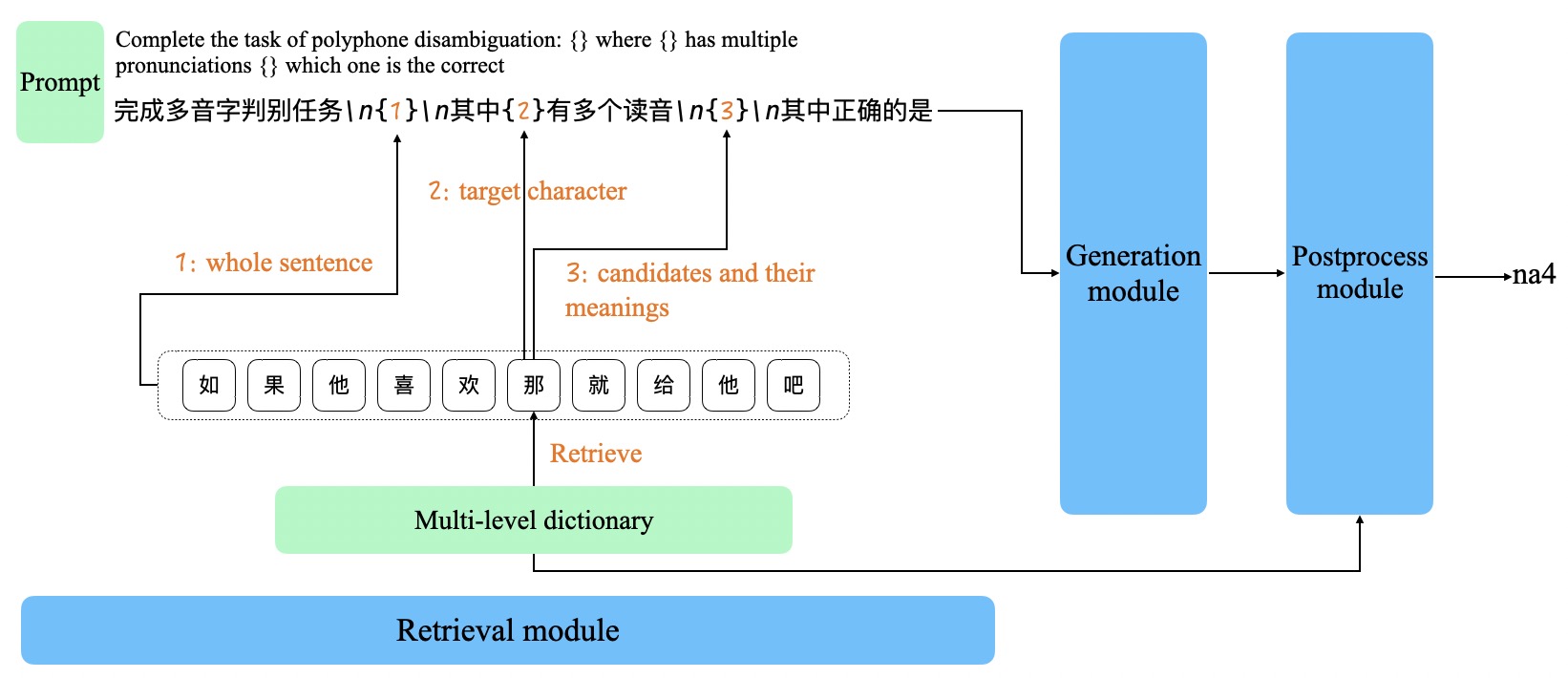}
    \caption{The overview of our proposed method}
    \label{fig:f2}
\end{figure}

Different from the existing classification-based methods, we apply generation-based LLM in the field of polyphone disambiguation. Figure \ref{fig:f2} shows the overall structure of the model, mainly consisting of three modules: Retrieval module, Generation module, and Postprocess module. We also augment the model with the help of a semantic dictionary for activating LLM abilities.

\subsection{Retrieval module}
\label{ssec:subhead}
As shown in Figure \ref{fig:f2}, this stage converts the sentence to the prompt which is the actual model input. The goal is to describe the task in natural language as much as possible to align with the pre-training process of LLM. The candidates and their meanings are retrieved to restrict the answer space based on a newly built semantic dictionary.

\subsubsection{Multi-level dictionary}
\label{sssec: multi-level dictionary}
\begin{figure}
    \centering
    \includegraphics[width=1\linewidth]{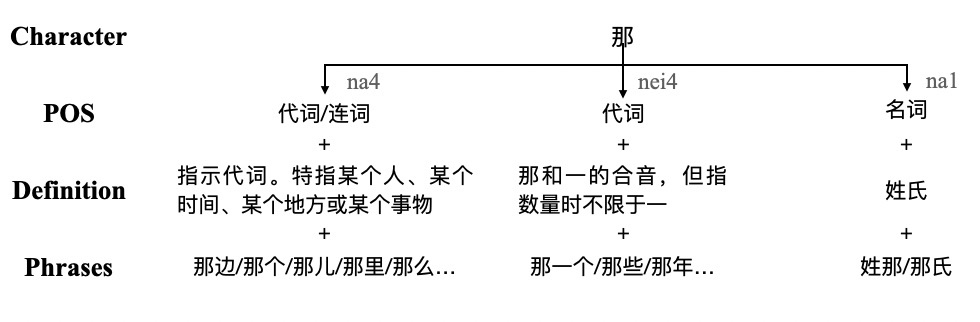}
    \caption{Samples of the multi-level dictionary}
    \label{fig:f3}
\end{figure}

The conventional methods only utilize the features of polyphonic characters and their context in sentences. We find that some external knowledge is also useful, such as the meaning and collocations of the characters. If people want to clarify the pronunciation of a polyphonic character, they usually look up a website dictionary and read the explanations to figure out which one is correct. Intuitively, we believe the information in the dictionary is beneficial for the disambiguation model. Therefore, based on the task, we construct a multi-level semantic dictionary from the Internet as shown in Figure \ref{fig:f3}. First, all characters that have more than one pinyin are extracted from the raw dictionary. Then the definitions and Part-of-Speech (POS) are bound with the corresponding pinyin as extensions. We also include the phrases containing these characters, in order to broaden the representations of each candidate pinyin. 

\subsubsection{Prompt}
\label{sssec:prompt}
\begin{figure}
    \centering
    \includegraphics[width=1\linewidth]{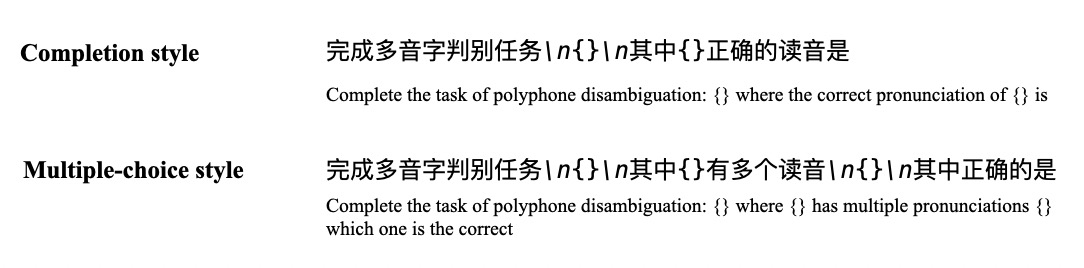}
    \caption{Different styles of prompts}
    \label{fig:f4}
\end{figure}

As indicated in Figure \ref{fig:f2}, Retrieval module produces a final prompt for the next step. The prompt-based method does the prediction tasks by utilizing language models that receive a template filled by the original input, and output the unfilled information of the template \cite{liu2023pre}. Our formatted natural language prompt consists of the task description, the input query, and/or external knowledge. We designed two types of prompts which are inspired by completion questions and multiple-choice questions. The details of these can be seen in Figure \ref{fig:f4}.

\subsection{Generation module}
\label{ssec:subhead}
\begin{figure}
    \centering
    \includegraphics[width=0.6\linewidth]{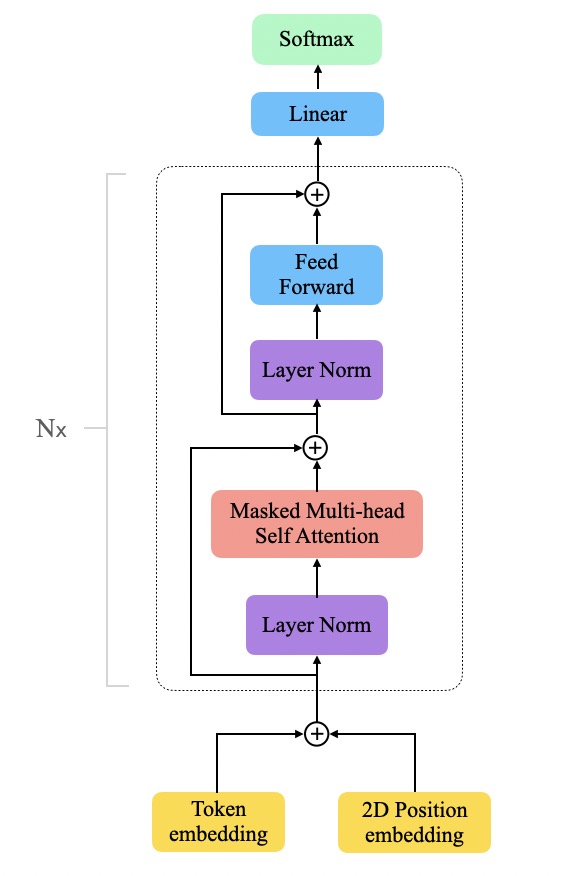}
    \caption{The modified Transformer architecture used in Generation module}
    \label{fig:f5}
\end{figure}

The existing LLMs can be roughly categorized into two major types, namely encoder-decoder and decoder-only \cite{zhao2023survey}. Our method selects decoder-only models to process the result of the previous module. The details are illustrated in Figure \ref{fig:f5}. Following the modifications of Transformer architecture mentioned in \cite{du2021glm}, the order of layer normalization and the residual connection is rearranged. A single linear layer is also added for the output token prediction. Besides, the activation functions, GeLUs, are replaced by ReLUs. To enhance the positional information among input tokens, a 2D positional encoding technique is applied which provides each token with two positional IDs. Specifically, the first id represents the position in the input sequence and the output tokens share the same position with the [MASK] token. The second id represents the intra-span position that ranges from 1 to the length of the answer span and the rest are 0. One advantage of this design is to fit downstream tasks as the length of the generated text is usually unknown beforehand.

\subsection{Postprocess module}
\label{ssec:subhead}
Due to the uncertainty of generation, the decoding model might output text that is not in valid candidates. For a well-tuned model, the proportion of this situation is quite small which is way less than 1\% according to our experiments. If it happens, we choose the candidate pinyin with the lowest edit distance from the prediction. When encountering the tie, we choose the first one which is also the more frequent one, cause the candidate list is sorted by frequency.

\section{Experiments}
\label{sec:experiments}

\subsection{Dataset}
\label{ssec:subhead}
To evaluate the proposed method, we did a few experiments on the CPP dataset (Chinese Polyphone with Pinyin) which is publicly available \cite{park2020g2pm}. The dataset consists of 99,264 sentences with lengths between 5 and 50 characters, which are collected from Chinese text in Wikipedia. Each of the sentences includes a polyphonic character marked by two special symbols (U+2582), and its corresponding correct pinyin annotation. The whole dataset is divided into train, development, and test sets with a ratio of 8:1:1. The datasets contain 623 polyphonic characters in total. Most of them have two possible pinyin (88.8\%) and the rest can have up to five pronunciations.

\subsection{Implementations}
\label{ssec:subhead}

\begin{CJK*}{UTF8}{gbsn}

\subsubsection{Baseline models}
\label{sssec:subhead}
To evaluate the proposed method, We implement five baseline models for comparison, and the details of these are listed as follows:
\begin{itemize}
    \item {\bf Majority vote}: It is a simple way to predict the target polyphone according to the statistics. We choose the most frequent pinyin of a polyphonic character in the training set as the result. For example, luo4 is picked for “咯” all the time regardless of the context while inference.
    \item {\bf G2Pm (BLSTM)}: A Bi-LSTM encoder accounts for learning the contextual information of a polyphonic character and then two fully connected layers are used to transform it into the classification label \cite{park2020g2pm}. The model utilizes cross-entropy as a loss function for training.
    \item {\bf G2Pm (BERT)}: Similar to the last model, we use the pre-trained Chinese BERT model \cite{cui2021pre} as the encoder of the contextual information. The output hidden state of the polyphonic character is fed to the fully connected layer to get the prediction. The whole model is trained in a fine-tuning manner with all weights tunable.
    \item {\bf PDF}: Following the model structure in \cite{zhang2021polyphone}, the model is composed of four parts: Input span feature layer,  PLM encoder layer, Transformer encoder Layer, and Restricted output Layer.
    \item {\bf g2pW}: Following the model structure in \cite{chen2022g2pw}, we implement the g2pW model for comparison. The model learns a soft-weighting function for the candidate pinyin in the Softmax. The function takes the outputs of BERT with the target polyphonic character and its Part-Of-Speech (POS) tagging. All hyper-parameters are the same as the original work.
\end{itemize}

\end{CJK*}

\subsubsection{Our model}
\label{sssec:subhead}
We apply a parameter-efficient fine-tuning method for our model based on P-tuning v2 \cite{liu2021p}. The parameters of the pre-trained language model are frozen and extra task-specific parameters are added to the model for training. First, according to the defined prefix length, a sequence of numbers is generated as a soft prompt. Then the soft prompt is encoded to parameter matrices that are trainable in different layers as illustrated in \cite{liu2021p}. In the training process, we choose ChatGLM-6B as the LLM backbone. We take a padding-to-right strategy to unify the input of one batch. The batch size and the prefix length are set to 32 and 64 respectively. AdamW is selected as the optimizer with a learning rate of 1e-02.

\subsection{Result and analysis}
\label{ssec:subhead}

\subsubsection{Performance of different models}
\label{sssec:subhead}
To evaluate the proposed method, we compared its performance with the above baseline models. Table \ref{tab:t1} lists the results of different models. As it shows, our model achieves an accuracy of 99.29\% and outperforms the other models. Compared with the state-of-the-art model, g2pW, our model has a 0.21\% test accuracy gain. It also surpasses G2Pm(BERT) and PDF with a 1.44\% and 0.46\% increase respectively.

\begin{table}[h]
\centering
\caption{Results on the CPP test set}
\label{tab:t1}
\begin{tabular}{lc}
\hline
\multicolumn{1}{c}{Model} & Test Acc. (\%) \\ \hline
Majority vote             & 92.08          \\
G2Pm (BLSTM)              & 97.31          \\
G2Pm (BERT)               & 97.85          \\
PDF                       & 98.83          \\
g2pW                      & 99.08          \\
Ours                      & \textbf{99.29} \\ \hline
\end{tabular}
\end{table}

\subsubsection{Effectiveness of external knowledge}
\label{sssec:subhead}
To verify the effectiveness of the multi-level dictionary, we further conduct experiments on whether to inject external knowledge into the prompt or not. As shown in Table \ref{tab:t3}, the model that takes the extra semantic information performs better than the other one with 0.34\% higher accuracy, proving that this feature is helpful for the task. It is also noticeable that the model without external knowledge underperforms g2pW slightly. However, it is not a fair comparison because g2pW used extra supervised information.

\begin{table}[h]
\centering
\caption{Results of varying models by the condition of external knowledge}
\label{tab:t3}
\begin{tabular}{lc}
\hline
\multicolumn{1}{c}{Model}  & Test Acc. (\%) \\ \hline
Without external knowledge & 98.95          \\
With external knowledge    & 99.29          \\ \hline
\end{tabular}
\end{table}

\subsubsection{Evaluation of different prompts}
\label{sssec:subhead}
As mentioned in section \ref{sssec:prompt}, the different designs of prompts would affect the model performance. Table \ref{tab:t2} presents the results of the two types of prompts. For a fair comparison, both prompts do not incorporate external knowledge. The model of the multiple-choice style achieves an accuracy improvement of 0.53\% from that of the completion style. It demonstrates that the provided candidates help the model restrict the output and understand the task better.

\begin{table}[h]
\centering
\caption{Results of different prompt types}
\label{tab:t2}
\begin{tabular}{lc}
\hline
\multicolumn{1}{c}{Prompt type} & Test Acc. (\%) \\ \hline
Completion style                & 98.42          \\
Multiple-choice style           & 98.95          \\ \hline
\end{tabular}
\end{table}

\subsubsection{Robustness to size of training data}
\label{sssec:subhead}
Table \ref{tab:t4} indicates that our model maintains competitive performance while the size of training data ranges. Our model also takes fewer training epochs (5 epochs) to achieve the best result, compared with PDF and g2pW.

\begin{table}[h]
\centering
\caption{Results of varying models by the size of training data}
\label{tab:t4}
\begin{tabular}{cc}
\hline
Ratio of used training data & Test Acc. (\%) \\ \hline
60\%                        & 98.84          \\
80\%                        & 99.06          \\
100\%                       & 99.29          \\ \hline
\end{tabular}
\end{table}

\begin{CJK*}{UTF8}{gbsn}

\subsubsection{Adaption to unseen characters}
\label{sssec:subhead}
Most of the existing methods are unable to predict unseen polyphonic characters, while our model naturally supports it. We take an example of the polyphonic character “红” that is not included in the CPP dataset. The character has two possible pinyin, \underline{hong2} and \underline{gong1}. The input sentence is “农夫释耒，红女下机” (Farmers drop plow and female workers quit spinning). Our model can correctly predict the result \underline{gong1}.

\end{CJK*}

\section{Conclusion}
\label{sec:print}
In this paper, we propose a novel polyphone disambiguation method by using LLM augmented by external knowledge. We construct a multi-level semantic dictionary for all polyphonic characters and incorporate it into the prompt. Experimental results show the effectiveness of the proposed method, thus outperforming the baseline models on the public CPP dataset. Some directions can be further studied in the future such as how the scale of LLM affects and how to introduce Chain-of-Thought techniques into the task.


\vfill\pagebreak

\bibliographystyle{IEEEbib}
\bibliography{strings,refs}

\end{document}